\documentclass[10pt,journal,compsoc]{IEEEtran}

\ifCLASSOPTIONcompsoc
  \usepackage[nocompress]{cite}
\else
  \usepackage{cite}
\fi

\ifCLASSINFOpdf
\else
\fi

\usepackage{dblfloatfix}
\usepackage{booktabs}
\usepackage{graphicx}

\usepackage{color}

\usepackage{xcolor}

\usepackage{makecell}

\usepackage[hyphens]{url}
\usepackage[hidelinks, pagebackref=true,breaklinks=true,colorlinks,bookmarks=false]{hyperref}
\hypersetup{breaklinks=true}
\begin{document}
%
\title{VIPriors 3: Visual Inductive Priors for\\Data-Efficient Deep Learning Challenges}
%
%
%
%

\author{Robert-Jan~Bruintjes,
        Attila~Lengyel,
        Marcos~Baptista~Rios,
        Osman~Semih~Kayhan,
        Davide~Zambrano,
        Nergis~Tomen
        and~Jan~van~Gemert
\IEEEcompsocitemizethanks{
\IEEEcompsocthanksitem R.J. Bruintjes, A. Lengyel, N. Tomen and J. van Gemert are with Delft University of Technology.\protect\\
E-mail: r.bruintjes@tudelft.nl
\IEEEcompsocthanksitem M. Baptista Rios is with Gradiant.
\IEEEcompsocthanksitem O. S. Kayhan is with Bosch Security Systems B.V.
\IEEEcompsocthanksitem D. Zambrano is with Synergy Sports.
}
}

%
%

\IEEEtitleabstractindextext{%
\begin{abstract}

The third edition of the "VIPriors: Visual Inductive Priors for Data-Efficient Deep Learning" workshop featured four data-impaired challenges, focusing on addressing the limitations of data availability in training deep learning models for computer vision tasks. The challenges comprised of four distinct data-impaired tasks, where participants were required to train models from scratch using a reduced number of training samples. The primary objective was to encourage novel approaches that incorporate relevant inductive biases to enhance the data efficiency of deep learning models. To foster creativity and exploration, participants were strictly prohibited from utilizing pre-trained checkpoints and other transfer learning techniques. Significant advancements were made compared to the provided baselines, where winning solutions surpassed the baselines by a considerable margin in all four tasks. These achievements were primarily attributed to the effective utilization of extensive data augmentation policies, model ensembling techniques, and the implementation of data-efficient training methods, including self-supervised representation learning. This report highlights the key aspects of the challenges and their outcomes.

\end{abstract}

\begin{IEEEkeywords}
Visual inductive priors, challenge, image classification, object detection, instance segmentation, action recognition.
\end{IEEEkeywords}}

\maketitle

\IEEEdisplaynontitleabstractindextext

%
\IEEEpeerreviewmaketitle


%
%
%
%

\IEEEraisesectionheading{\section{Introduction}\label{sec:introduction}}

\IEEEPARstart{D}{ata} is fueling deep learning, yet obtaining high quality annotations is often costly. In recent years, extensive research has been dedicated to exploring ways to utilize large quantities of data to train comprehensive foundation models for vision and language~\cite{Bommasani2021OnTO}, and to combine multiple modalities for weak supervision~\cite{radford21learning}. While these approaches have demonstrated impressive results, self-supervision is not yet the holy grail. Training on massive datasets still requires a significant amount of energy, contributing to carbon emissions. Furthermore, only a handful of deep learning behemoths have access to billions of data points and expensive deep learning hardware. In addition, large quantities of data may simply not be available in certain domains. The Visual Inductive Priors for Data-Efficient Deep Learning workshop (VIPriors) therefore aims to encourage research on learning efficiently from few data samples by combining the power of deep learning with hard-won knowledge priors from various fields. We focus on data efficiency through visual inductive priors.

The Visual Inductive Priors for Data-Efficient Deep Learning workshop has now been organized for the third year in a row, with the latest 2022 edition taking place at ECCV in Tel Aviv, Israel. In order to stimulate research in data-efficient computer vision, the workshop includes challenges in which participants train computer vision models on small subsets of (publicly available) datasets. We challenge participants to submit solutions that are able to learn an effective representation of the dataset without access to the large quantities of data that is used to train state-of-the-art deep computer vision models.

In this report, we present the outcomes of the third edition of these challenges. We discuss specific details and top-ranking solutions of each challenge. It was observed that the top competitors in all challenges heavily relied on model ensembling and data augmentation to improve the data efficiency of their solutions. Additionally, many of the participants' solutions utilized a limited number of backbones and baseline methods, which seem to possess properties conducive to learning from small data. To recognize submissions that introduce innovative methods, a jury prize for each challenge was awarded to the most noteworthy submission.

\begin{table*}[ph]
\centering
\caption{Overview of challenge submissions. J indicates jury prize. Bold-faced methods are contributions by the competitors.}
\label{tab:conclusion}
\renewcommand{\arraystretch}{1.4}
\scalebox{0.9}{
\begin{tabular}{@{}lllllc@{}}
\toprule
Ranking & Teams & Encoder architectures & Data augmentation & Methods & Main metric \\ \midrule

\multicolumn{2}{@{}l}{\textbf{Classification}} & & & \\ \midrule
1 & \makecell[l]{\textbf{Ma et al.}} & \makecell[l]{SE+PyramidNet, ResNeSt200e,\\ReXNet,EfficientNet-B8,\\ConvNeXt-XL*} & \makecell[l]{CutMix, AutoAugment~\cite{cubuk2018autoaugment}, \\Stubborn Image\\Augmentation(SIA)} & \makecell[l]{Label smoothing, AdvProp,\\Random image cropping and patching\\ (RICP), extra training on \\stubborn images, hard fusion} & \textbf{78.7} \\
2 \& J & \makecell[l]{Lu et al.} & \makecell[l]{HorNet, ConvNeXt} & \makecell[l]{Automix \cite{liu2021unveiling}} & \makecell[l]{Cross-decoupled\\ knowledge distillation \cite{zhao2022decoupled},\\ label smoothing} & 77.9 \\
3 & \makecell[l]{Zuo et al.} & \makecell[l]{CoAtNet, TResNet, Resnet50,\\ Resnext50, EdgeNeXt} & \makecell[l]{CutMix, Random erasing, \\MixUp, AutoAugment} & \makecell[l]{Knowledge distillation between encoders} & 77.7 \\
4 & \makecell[l]{Wang et al.} & \makecell[l]{ResNeSt, TResNet, SE-ResNet, \\ReXNet, ECA-NFNet, ResNet-RS \cite{bello2021revisiting},\\ Inception-ResNet, RegNet, \\EfficientNet, MixNet} & \makecell[l]{AutoAugment, MixUp, CutMix,\\ padding} & \makecell[l]{label smoothing, \\train on larger images, \\data resampling} & 76.8 \\
5 & \makecell[l]{She et al.} & \makecell[l]{ResNeSt, Res2Net, Xception,\\DPN \cite{chen2017dual}, EfficientNet, SENet} & \makecell[l]{AutoAugment, MixUp} & \makecell[l]{label smoothing, \\train on larger images,\\ hard negative resampling} & 75.4 \\
6 & \makecell[l]{Chen et al.} & \makecell[l]{ResNeSt, EfficientNet, ReXNet,\\ RegNetY} & \makecell[l]{AutoAugment, MixUp, \\CutMix, ColorJitter} & \makecell[l]{label smoothing, \\train on larger images, \\Exponential Moving Average on\\ network parameters} & 70.8 \\
\midrule

\multicolumn{2}{@{}l}{\textbf{Object detection}} & & & \\ \midrule
1 & \makecell[l]{\textbf{Lu et al.}} & \makecell[l]{YOLOv4\cite{bochkovskiy2020yolov4}, YOLOv7\cite{ yolov7}, \\ YOLOR~\cite{yolor}, CBNetv2~\cite{cbnetv2}}  & \makecell[l]{Mosaic~\cite{bochkovskiy2020yolov4}, mix-up~\cite{zhang2018mixup}, \\ copy-paste~\cite{kisantal2019augmentation}} & 
\makecell[l]{Weighted Boxes Fusion~\cite{solovyev2021weighted},\\ TTA, Model Soups~\cite{wortsman2022model}, \\ Image Uncertainty Weighted} & \textbf{33.0}       \\ \\

2 & \makecell[l]{Xu et al.} & \makecell[l]{Cascade RCNN~\cite{cai2018cascade}, Swin T.~\cite{liu2021Swin}, \\ ConvNext~\cite{liu2022convnet}, ResNext~\cite{xie2017aggregated}} & 
\makecell[l]{AutoAugment~\cite{cubuk2018autoaugment}, \\ random flip, \\ multi-scale augmentations~\cite{liu2000multi}} & \makecell[l]{MoCoV3~\cite{chen2021empirical}, MoBY~\cite{xie2021self}, \\ Soft-NMS~\cite{bodla2017soft},  FPN~\cite{lin2017feature}, SSFPN\cite{hong2021sspnet}, \\ non-maximum weighted (NMW)~\cite{zhou2017cad}} & 32.9          \\ \\

3 & \makecell[l]{J. Zhao et al.} &  \makecell[l]{Cascade RCNN~\cite{cai2018cascade}, Swin T.~\cite{liu2021Swin}, \\ Convnext~\cite{liu2022convnet}} & \makecell[l]{Albu, MixUp~\cite{zhang2018mixup}, \\ AutoAugment~\cite{cubuk2018autoaugment}}  & \makecell[l]{Stochastic Weight Averaging~\cite{izmailov2018averaging}, \\ Hard classes retraining, \\ FPN~\cite{lin2017feature}, Soft-NMS~\cite{bodla2017soft}, \\ pseudo labeling}  & 32.4        \\ \\

4 \& J  & \makecell[l]{P. Zhao et al.} & \makecell[l]{Cascade RCNN~\cite{cai2018cascade}, Swin T.~\cite{liu2021Swin}, \\ Pyramid ViT~\cite{wang2021pyramid}} & \makecell[l]{Mosaic~\cite{bochkovskiy2020yolov4},  MixUp~\cite{zhang2018mixup}} &  \makecell[l]{SimMIM~\cite{xie2022simmim},  GIOU loss~\cite{union2019metric}, \\ Soft-NMS~\cite{bodla2017soft}} & 30.9         \\ \\ 
\midrule


\multicolumn{2}{@{}l}{\textbf{Instance segmentation}} & & & \\ \midrule

1 & \makecell[l]{\textbf{Yan et al.}~\cite{yan1seg}} &
\makecell[l]{CBSwin-T~\cite{liang2021cbnetv2}} &
\makecell[l]{\textbf{TS-DA}, \textbf{TS-IP}~\cite{yan1seg} \\ Random scaling, cropping} &
\makecell[l]{Hybrid Task Cascade~\cite{Chen_2019_CVPR}, \\ CBFPN~\cite{liang2021cbnetv2}} &
\textbf{53.1} \\

2 (shared) & \makecell[l]{Leng et al.} &
\makecell[l]{Swin Transformer-Large~\cite{liu2021Swin}} &
\makecell[l]{AutoAugment~\cite{cubuk2018autoaugment}, \\ 
ImgAug~\cite{imgaug}, Copy-Paste~\cite{ghiasi2021copypaste}, \\ Horizontal Flip and \\ Multi-scale Training} &
\makecell[l]{CBNetV2~\cite{liang2021cbnetv2}}
& 50.6 \\

 & \makecell[l]{Lu et al.} &
\makecell[l]{CBSwin-T~\cite{liang2021cbnetv2} \\ ResNet~\cite{he2015deep} \\ ConvNeXt~\cite{liu2022convnet} \\ Swinv2~\cite{liu2021Swin} \\ CBNetv2~\cite{liang2021cbnetv2}} &
\makecell[l]{MixUp~\cite{zhang2018mixup}, Mosaic \\ Task-Specific Copy-Paste~\cite{yunusov2021instance} \\Color and geometric \\ transformations} &
\makecell[l]{Hybrid Task Cascade~\cite{Chen_2019_CVPR} \\ CBFPN~\cite{liang2021cbnetv2} \\ Group Normalization~\cite{Wu2018GroupN}}
& 50.6 \\

3 & \makecell[l]{Zhang et al.} & \makecell[l]{CBSwin-T~\cite{liang2021cbnetv2}} &
\makecell[l]{Location-aware MixUp, \\ RandAugment~\cite{cubuk2020randaugment}, \\ GridMask~\cite{chen2020gridmask}, \\ Random scaling, \\  Copy-paste~\cite{ghiasi2021copypaste}, \\ Multi-scale augmentation, \\ TTA} &
\makecell[l]{Hybrid Task Cascade~\cite{Chen_2019_CVPR} \\ Seesaw Loss~\cite{Wang_2021_CVPR} \\ SWA~\cite{Izmailov2019averaging}} &
49.8 \\

4 & \makecell[l]{Cheng et al.} & CBSwin-T~\cite{liang2021cbnetv2} &
\makecell[l]{RandAugment \\ Copy-Paste~\cite{ghiasi2021copypaste} \\ GridMask} &
\makecell[l]{Hybrid Task Cascade~\cite{Chen_2019_CVPR} \\ Mask Transfiner~\cite{transfiner}} &
18.5 \\

5 \& J & \makecell[l]{Cheng et al.~\cite{Cheng2022SparseInst}} & ResNet~\cite{he2015deep} &
\makecell[l]{Random flip and scale jitter} &
\makecell[l]{Sparse Instance Activation \\ for Real-Time Instance \\ Segmentation~\cite{Cheng2022SparseInst}} &
18.5 \\ \midrule


\multicolumn{2}{@{}l}{\textbf{Action recognition}} & & & \\ \midrule

1 &
\makecell[l]{\textbf{Song et al.}} &
\makecell[l]{
    R(2+1)D~\cite{Tran2018r21d}, SlowFast~\cite{Feichtenhofer2019sf}, \\
    CSN~\cite{csn}, X3D~\cite{x3d}, TANet~\cite{liu2019tanet}, \\
    Timesformer~\cite{timesformer}
} & 
Random flipping, TenCrop &
Soft voting &
0.71 \\

2 &
\makecell[l]{He et al.} &
\makecell[l]{SlowFast~\cite{Feichtenhofer2019sf}, Timesformer~\cite{timesformer}, \\ TIN~\cite{shao2020temporal}, TPN~\cite{tpn}, X3D~\cite{x3d}, \\
Video Swin Transformers~\cite{liu2021video}, \\ R(2+1)D~\cite{Tran2018r21d}, DirecFormer~\cite{truong2021direcformer}.} &
\makecell[l]{AutoAugment~\cite{cubuk2018autoaugment}, CutMix~\cite{yun2019cutmix} \\ random flip, grayscale, jitter, \\ temporal aug., TenCrop, \\ test-time aug.} &
Label smoothing &
0.69 \\

3 \& J &
\makecell[l]{Tan et al.} &
\makecell[l]{TSN~\cite{wang2016temporal}, TANet~\cite{liu2019tanet}, TPN~\cite{tpn}, \\ SlowFast~\cite{Feichtenhofer2019sf}, CSN~\cite{csn}, \\ Video MAE~\cite{tong2022videomae}} &
MixUp~\cite{zhang2018mixup}, CutMix~\cite{yun2019cutmix} &
MoCo~\cite{he2020momentum}, TVL-1~\cite{zach2017aduality} &
0.59 \\
\bottomrule
\end{tabular}
}
\end{table*}
\section{Challenges}
\label{sec:challenges}

The workshop accommodates four common computer vision challenges in which the number of training samples are reduced to a small fraction of the full set:

\textbf{Image classification}: We use a subset of Imagenet~\cite{deng2009imagenet}. The subset contains 50 images from 1,000 classes for training, validation and testing.

 \textbf{Object detection}: DelftBikes~\cite{kayhan2021hallucination} dataset is used for the object detection challenge. The dataset includes 8,000 bike images for training and 2,000 images for testing (Fig.~\ref{fig:bike_images}). Each image contains 22 different bike parts that are annotated as bounding box, class and object state labels.

\textbf{Instance segmentation}: The main objective of the challenge is to segment basketball players and the ball on images recorded of a basketball court. The dataset is provided by SynergySports\footnote{\url{https://synergysports.com}} and contains a train, validation and test set of basketball games recorded at different courts with instance labels.

\textbf{Action recognition}: For this challenge we have provided Kinetics400ViPriors, which is an adaptation of the well-known Kinetics400 dataset~\cite{kinetics400}. The training set consists of approximately 40k clips, while the validation and test sets contain about 10k and 20k clips, respectively.

We provide a toolkit\footnote{\url{https://github.com/VIPriors/vipriors-challenges-toolkit}} which consists of guidelines, baseline models and datasets for each challenge.
The competitions are hosted on the Codalab platform. Each participating team submits their predictions computed over a test set of samples for which labels are withheld from competitors.

The challenges include certain rules to follow:
\begin{itemize}
    \item Models ought to train from scratch with only the given dataset.
    \item The usage of other data rather than the provided training data, pretraining the models and transfer learning methods are prohibited. 
    \item The participating teams need to write a technical report about their methodology and experiments.
\end{itemize}

\textbf{Shared rankings.} Due to confusion around the exact deadline of the competitions, we have merged rankings of two different moments. This has resulted in shared places in some of the rankings of the individual challenges.
\subsection{Classification}

Image classification serves as an important benchmark for the progress of deep computer vision research. In particular, the ImageNet dataset~\cite{deng2009imagenet} has been the go-to benchmark for image classification research. ImageNet gained popularity because of its significantly larger scale than those of existing benchmarks. Ever since, even larger datasets have been used to improve computer vision, such as the Google-owned JFT-300M~\cite{sun2017revisiting}. However, we anticipate that relying on the increasing scale of datasets is problematic, as increased data collection is expensive and can clash with privacy interests of the subjects. In addition, for domains like medical imaging, the amount of labeled data is limited and the collection and annotation of such data relies on domain expertise. Therefore, we posit that the design of data efficient methods for deep computer vision is crucial. 

As in the two earlier editions of this workshop, in our image classification challenge we provide a subset~\cite{kayhan2020translation} of the Imagenet dataset~\cite{deng2009imagenet} consisting of 50 images per class for each of the train, validation and test splits. 
The classification challenge had 14 participating teams, of which six teams submitted a report. The final ranking and the results can be seen in Table~\ref{tab:classification}.

\begin{table*}[t]
\centering
\caption{Final rankings of the Image Classification challenge.}
\renewcommand{\arraystretch}{2.0}
\begin{tabular}{@{}llc@{}}
\toprule
Ranking & Teams                & Top-1 Accuracy \\ \midrule
1 & \makecell[l]{\textbf{Tianzhi Ma, Zihan Gao, Wenxin He, Licheng Jiao} \\ \textbf{\textit{School of Artificial Intelligence, Xidian University.}}} & \textbf{78.7} \\
2 \& J & \makecell[l]{Xiaoqiang Lu, Chao Li, Chenghui Li, Xiao Tan, Zhongjian Huang, Yuting Yang \\ \textit{School of Artificial Intelligence, Xidian University.}} & 77.9 \\
3 & \makecell[l]{Yi Zuo, Zitao Wang, Xiaowen Zhang, Licheng Jiao \\ \textit{School of Artificial Intelligence, Xidian University.}} & 77.7 \\
4 & \makecell[l]{Jiahao Wang, Hao Wang, Hua Yang, Fang liu, Lichang Jiao \\ \textit{School of Artificial Intelligence, Xidian University.}} & 76.8 \\
5 & \makecell[l]{Wenxuan She, Mengjia Wang, Zixiao Zhang, Fang Liu, Licheng Jiao \\ \textit{School of Artificial Intelligence, Xidian University.}} & 75.4 \\
6 & \makecell[l]{Baoliang Chen, Yuxuan Zhao, Fang Liu, Licheng Jiao \\ \textit{School of Artificial Intelligence, Xidian University.}} & 70.8 \\

\bottomrule
\end{tabular}
\label{tab:classification}
\end{table*}
\subsubsection{First place}

The team from Xidian University led by Tianzhi Ma uses an ensemble of SE+PyramidNet, ResNeSt200e, ReXNet, EfficientNet-B8 and ConvNeXt-XL* models. They apply diverse data augmentation strategies to increase the diversity in the data, and include several other optimization tricks like RICP and hard fusion. Ultimately, they were able to improve their top-1 accuracy from last year's challenge (68.6) to a winning top-1 accuracy of 79\%. 


\subsubsection{Second place \& jury prize}


The team from Xidian University led by Xiaoqiang Lu uses only two models in their ensemble, and instead gain performance by using cross-decoupled knowledge distillation \cite{zhao2022decoupled}. Other than this, only Automix \cite{liu2021unveiling} and label smoothing are required to secure second place. For this minimal yet effective solution we award this team the jury prize.

\subsubsection{Third place}


The team from Xidian University lead by Yi Zou uses five different encoder architectures. They exhaustively apply knowledge distillation from all encoders to all other encoders to train twenty models, which are all ensembled for the final model. All models are trained with severe data augmentation: CutMix, random erasing, MixUp, AutoAugment.

\subsubsection{Conclusion}

As in previous editions \cite{bruintjes2021vipriors, lengyel2022vipriors}, the crucial components of a competitive submission to the image classification competition are ensembling of many different classification architectures, as well as combining multiple different augmentation policies. Aside from label smoothing and training with larger image sizes, knowledge distillation gained in popularity among the methods used to train the networks.
\subsection{Object Detection}
\begin{figure}[!ht]
	\centering
	\includegraphics[width=\linewidth]{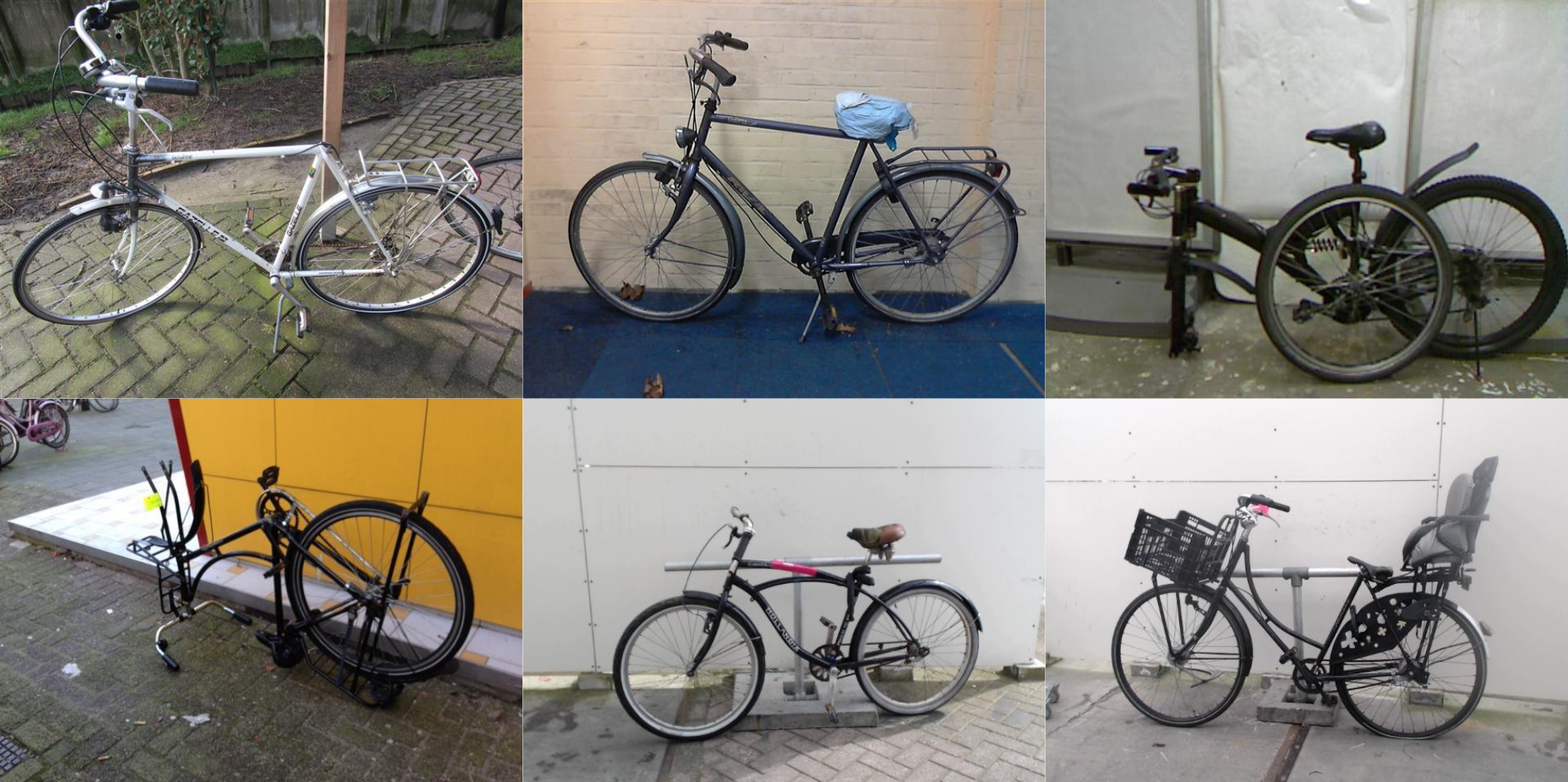}	
	\caption{Some images from the DelftBikes dataset. Each image has a single bike with 22 labeled parts. }
	\label{fig:bike_images}
\end{figure}

Similar to the object detection challenge last year~\cite{lengyel2022vipriors}, we also use DelftBikes~\cite{kayhan2021hallucination} dataset this year (Fig.~\ref{fig:bike_images}). Each image in the DelftBikes contains 22 labeled bike parts as class and bounding box labels of each part. In addition, the dataset includes extra object state labels as intact, missing, broken or occluded.
The dataset has 10k bike images in total and 2k of the images are used for only testing purposes.
The dataset contains different object sizes, and contextual and location biases that can cause false positive detections~\cite{kayhan2021hallucination, kayhan2022evaluating}. Note that, some of the object boxes are noisy which introduces more challenges to detect object parts.

We provide a baseline detector as a Faster RCNN with a Resnet-50 FPN~\cite{ren2016faster} backbone from scratch for 16 epochs. This baseline network is trained with the original image size without any data augmentation. It performs 25.8\% AP score on the test set.
Note that, the evaluation is done on available parts which are intact, damaged and occluded parts.


\begin{table*}[t]
\centering
\caption{Final rankings of the Image Object Detection challenge.}
\renewcommand{\arraystretch}{2.0}
\begin{tabular}{@{}llc@{}}
\toprule
Ranking & Teams & AP @ 0.5:0.95 \\ \midrule
1 & \makecell[l]{\textbf{Xiaoqiang Lu, Yuting Yang, Zhongjian Huang, Xiao Tan, Chenghui Li.} \\ \textbf{\textit{School of Artificial Intelligence, Xidian University}}} & \textbf{33} \\
2 & \makecell[l]{Bocheng Xu, Rui Zhang, and Yanyi Feng. \\ \textit{Department of AI R\&D, Terminus Technologies.}} & 32.9 \\
3 & \makecell[l]{Jiawei Zhao, Zhaolin Cui, Xuede Li, Xingyue Chen, Junfeng Luo, and Xiaolin Wei. \\ \textit{Vision Intelligence Department (VID).}}  & 32.1 \\
4 \& J & \makecell[l]{Ping Zhao, Xinyan Zhang, Weijian Sun, and Xin Zhang. \\ \textit{Huawei Technologies Co., Ltd. and Tongji University}} & 30.9         \\ 
 
\bottomrule
\end{tabular}
\label{tab:detection2}
\end{table*}

The detection challenge had 41 participant teams. The team from Xidian University obtained first place by 33\% AP scores. Terminus Technologies and Vision Intelligence Department from Meituan followed them by 32.9\% AP and 32.1\% AP respectively.
The team from Huawei Technologies and Tongji University won the jury prize for their 'coarse-to-fine' idea.


\subsubsection{First place}
\begin{figure}[!ht]
	\centering
	\includegraphics[width=\linewidth]{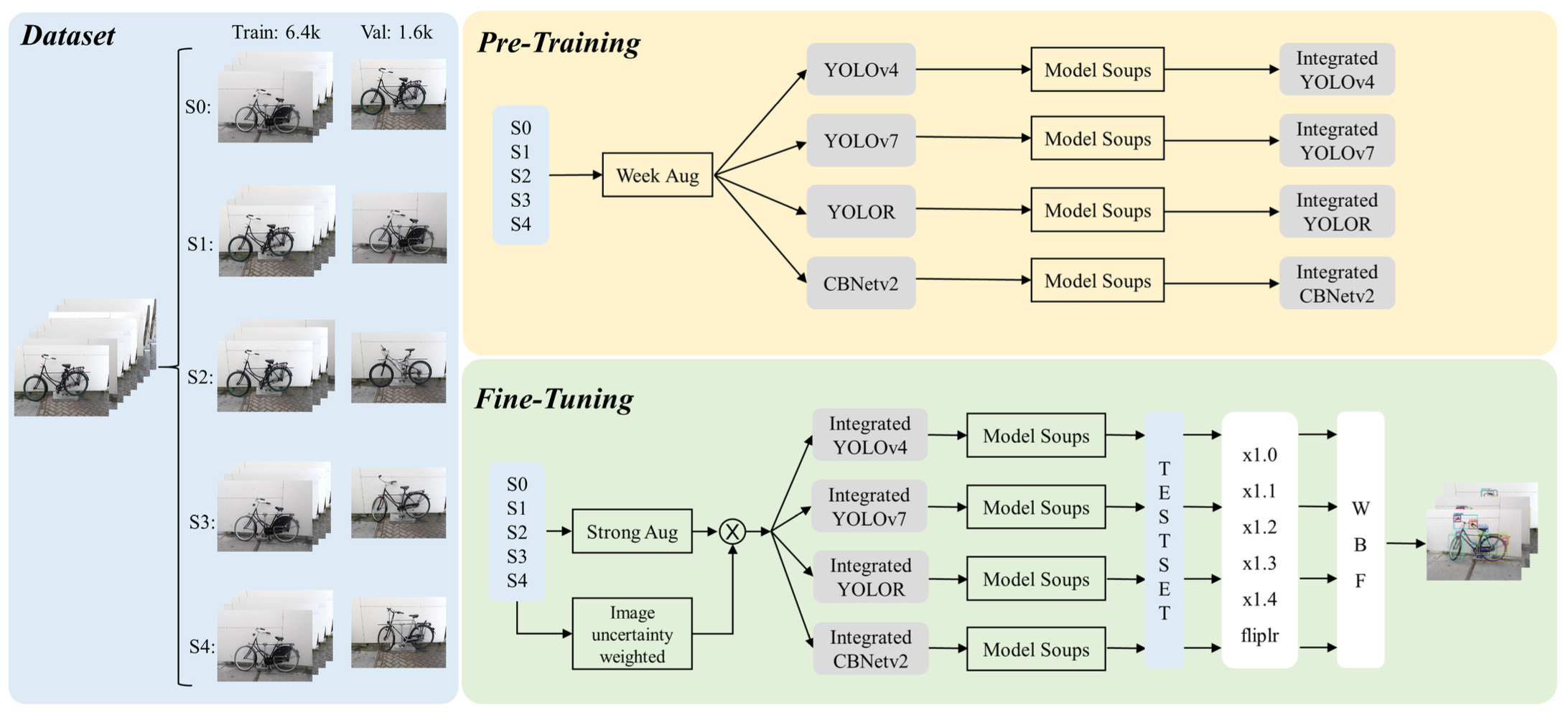}	
	\caption{Bag of Freebies for training detector~\cite{zhang2019bag}. They train different models during the pretraining and fine-tuning phases with different types of data augmentation methods. They also use image uncertainty to improve object detection performance.}
	\label{fig:arch11}
\end{figure}

Lu et al. employ an ensemble of various YOLO detectors~\cite{bochkovskiy2020yolov4, yolov7, yolor} and CBNetv2~\cite{cbnetv2} (Fig.~\ref{fig:arch11}). They design two-stage training: (i) pre-training by using weak data augmentation and (ii) fine-tuning  by using strong data augmentation such as mosaic~\cite{bochkovskiy2020yolov4}, mix-up~\cite{zhang2018mixup}, and copy-paste~\cite{kisantal2019augmentation} and a weighted training strategy based on image uncertainty. The authors further improved the results by weighted boxes fusion (WBF) ~\cite{solovyev2021weighted} and TTA strategies and obtain 33 \% AP on the test set.

\subsubsection{Second place}
Xu et al. train Cascade RCNN~\cite{cai2018cascade} using Swin Transformer~\cite{liu2021Swin}, ConvNext~\cite{liu2022convnet} and ResNext~\cite{xie2017aggregated} as backbone architectures. These backbones are pretrained by using self-supervised methods such as  MoCoV3~\cite{chen2021empirical} and MoBY~\cite{xie2021self}. In addition, they use AutoAugment~\cite{cubuk2018autoaugment}, random flip and multi-scale augmentation methods~\cite{liu2000multi} to improve the detection performance. Finally, the non-maximum weighted (NMW) ~\cite{zhou2017cad} method, Soft-NMS~\cite{bodla2017soft} and model ensemble methods are used on the test set. The method obtained 32.9\% AP on the test set.


\subsubsection{Third place}
\begin{figure}[!ht]
	\centering
	\includegraphics[width=\linewidth]{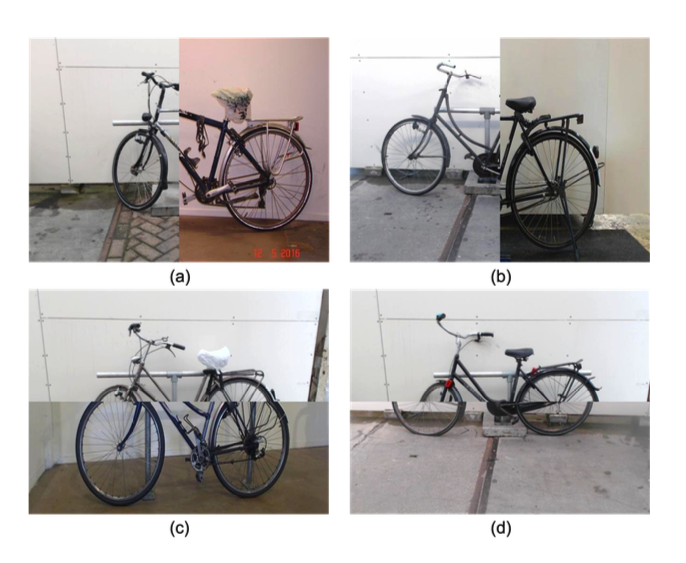}	
	\caption{Synthetic images generated for backbone pretraining.}
	\label{fig:synthetic}
\end{figure}

Zhao et al. initially train Cascade RCNN~\cite{cai2018cascade} detector using ConvNext backbone~\cite{liu2022convnet}. Then, they create a synthetic dataset (Fig.~\ref{fig:synthetic}) from the training set and obtain pseudo labels on this dataset with the initial trained model. Afterwards, they train the same model only on the pseudo labels with smaller-resolution images. In the end, they retrain the pseudo-label pretrained network with the original train set and select some hard classes to improve the detector performance on them. During training phases, they also use various data augmentation methods as colour jittering and RGB shifting, mix-up~\cite{zhang2018mixup} and AutoAugment~\cite{cubuk2018autoaugment}. The method obtains 32.1\% AP detection performance.

\subsubsection{Jury prize}
\begin{figure}[!ht]
	\centering
	\includegraphics[width=\linewidth]{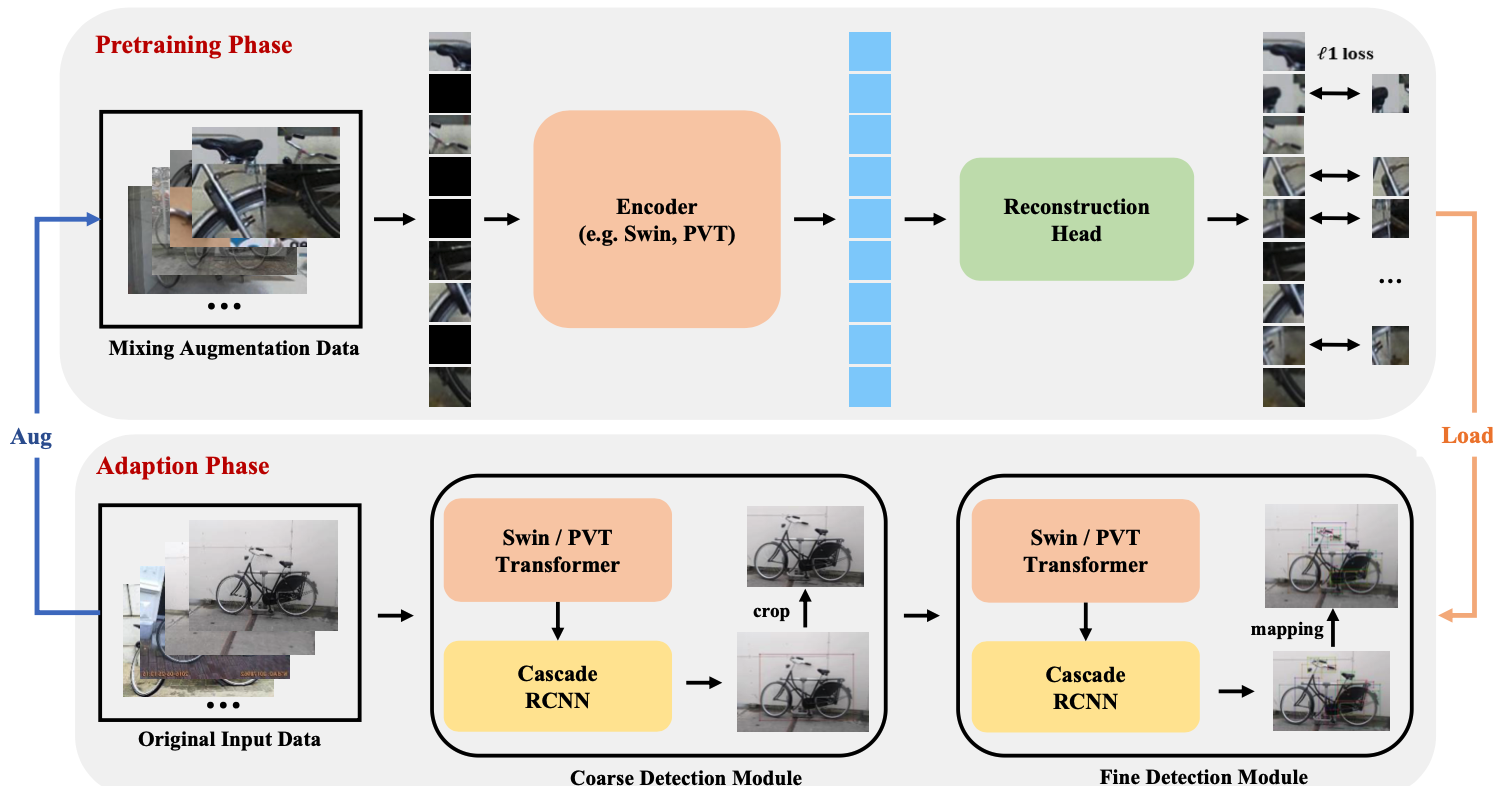}	
	\caption{Method pipeline. First, the backbone is pretrained by using SimMIM~\cite{xie2022simmim} to obtain strong features. In the adaptation phase, coarse to fine detection strategy improves detection.}
	\label{fig:coarse}
\end{figure}

Method of Zhao et al. has two phases: pretraining and adaptation phases (Fig.~\ref{fig:coarse}). In the pretraining phase, they utilize mosaic~\cite{bochkovskiy2020yolov4} and mix-up~\cite{zhang2018mixup} data augmentations on object and image level features and train  SimMIM~\cite{xie2022simmim}. In the adaptation phase, a pretrained encoder of SimMIM is used to initialize the backbone of Cascade RCNN~\cite{cai2018cascade}. In 'coarse detection', the model detects bike objects. In the 'fine detection' phase, the fine detection module runs on the cropped bike object from the previous phase and tries to detect relevant bike parts. The final model obtains 30.94\% AP. The team earned the jury prize because of their 'coarse-to-fine' idea, well-written article and discussion of strategies that did not work.

\subsection{Instance Segmentation}
Instance segmentation is the task of detecting and segmenting specific objects and instances in an image. With applications ranging from autonomous driving, surveillance, remote sensing to sport analysis, it is a fundamental computer vision problem. Similarly to last year, our challenge is based on the basketball dataset provided by SynergySports \cite{sportradar_dataset}, consisting of images recorded during various basketball games played on different courts. The goal is to detect and predict segmentation masks of all players and ball objects in the images. With a mere 184, 62, and 64 samples for the train, validation and test splits, respectively, the dataset is considered very small. The test labels are withheld from the challenge participants and final performance on the test set is evaluated on an online server. The main metric used is the Average Precision (AP) @ 0.50:0.95. Our baseline method is based on the Detectron2 \cite{wu2019detectron2} implementation of Mask-RCNN \cite{he2017mask}.

Twelve teams submitted solutions to the evaluation server, of which six teams submitted a report to qualify their submission to the challenge. The final rankings are shown in Table \ref{tab:segmentation}.

\begin{table*}[t]
\centering
\caption{Final rankings of the Instance Segmentation challenge. J indicates jury prize.}
\renewcommand{\arraystretch}{2.0}
\begin{tabular}[t]{@{}llc@{}}
\toprule
Ranking & Teams & \% AP @ 0.50:0.95 \\ \midrule

1 & \makecell[l]{
    \textbf{Bo Yan, Xingran Zhao, Yadong Li, Hongbin Wang.} \\
    \textbf{\textit{Ant Group, China.}}
    } & \textbf{53.1} \\
2 (shared) & \makecell[l]{
    Fuxing Leng, Jinghua Yan, Peibin Chen, Chenglong Yi. \\
    \textit{ByteDance, Huazhong University of Science and Technology.}} & 50.6 \\
    
  & \makecell[l]{
    Xiaoqiang Lu, Yuting Yang, Zhongjian Huang. \\
    \textit{School of Artificial Intelligence, Xidian University, Xi'an, China.}} & 50.6 \\

3 & \makecell[l]{
    Junpei Zhang, Kexin Zhang, Rui Peng, Yanbiao Ma, Licheng Jiao Fang Liu. \\
    \textit{Team Yanbiao\_Ma.}} & 49.8 \\ 

4 & \makecell[l]{
    Yi Cheng, ShuHan Wang, Yifei Chen, Zhongjian Huang. \\
    \textit{School of Artificial Intelligence, Xidian University, Xi'an, China.}} & 47.6 \\ 
    
5 \& J & \makecell[l]{
    Tianheng Cheng, Xinggang Wang, Shaoyu Chen, Qian Zhang, Chang Huang, Zhaoxiang Zhang, \\
    Wenqiang Zhang, Wenyu Liu. \\
    \textit{(1) School of EIC, Huazhong University of Science \& Technology; (2) Horizon Robotics;} \\
    \textit{(3) Institute of Automation, Chinese Academy of Sciences (CASIA)}} & 34.0 \\ \bottomrule
\end{tabular}
\label{tab:segmentation}
\end{table*}

\subsubsection{First place}
\begin{figure}
    \centering
    \includegraphics[width=\linewidth]{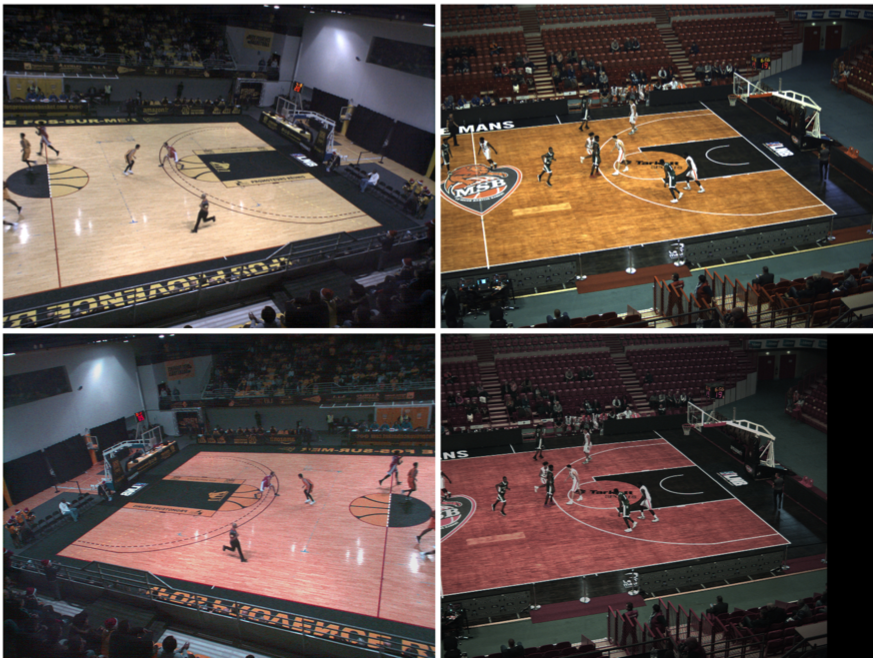}
    \caption{Data augmentation policy of the first-place instance segmentation submission by Yan et al.~\cite{yan1seg}.}
    \label{fig:3_1_seg}
\end{figure}

The method of Yan et al.~\cite{yan1seg} introduces a task-specific data augmentation (TS-DA) strategy to generate additional training data, and a task-specific inference processing (TS-IP) which is applied at test time. TS-DA employs Copy-Paste~\cite{ghiasi2021copypaste} augmentations with constraints on the absolute locations of the synthetic players and ball objects to ensure all objects are placed inside the court, and their relative locations to mimic player-ball interactions. Subsequently, geometric and photometric augmentations are applied to the image to further increase the variety in their appearance. During inference, random scaling and cropping is applied to the images, and additional filtering employed to the predictions to ensure only one basketball of reasonable dimensions is present on the court. The complete data augmentation policy is illustrated in Figure~\ref{fig:3_1_seg}.

The segmentation model is based on the Hybrid Task Cascade (HTC) detector~\cite{Chen_2019_CVPR} and the CBSwin-T backbone with CBFPN~\cite{liang2021cbnetv2}. Mask Scoring R-CNN~\cite{huang2019mask} is used to further improve segmentation quality. After training the model, it is further finetuned using the SWA~\cite{zhang2020swa} strategy.

\subsubsection{Shared second place - A}
\begin{figure}
    \centering
    \includegraphics[width=\linewidth]{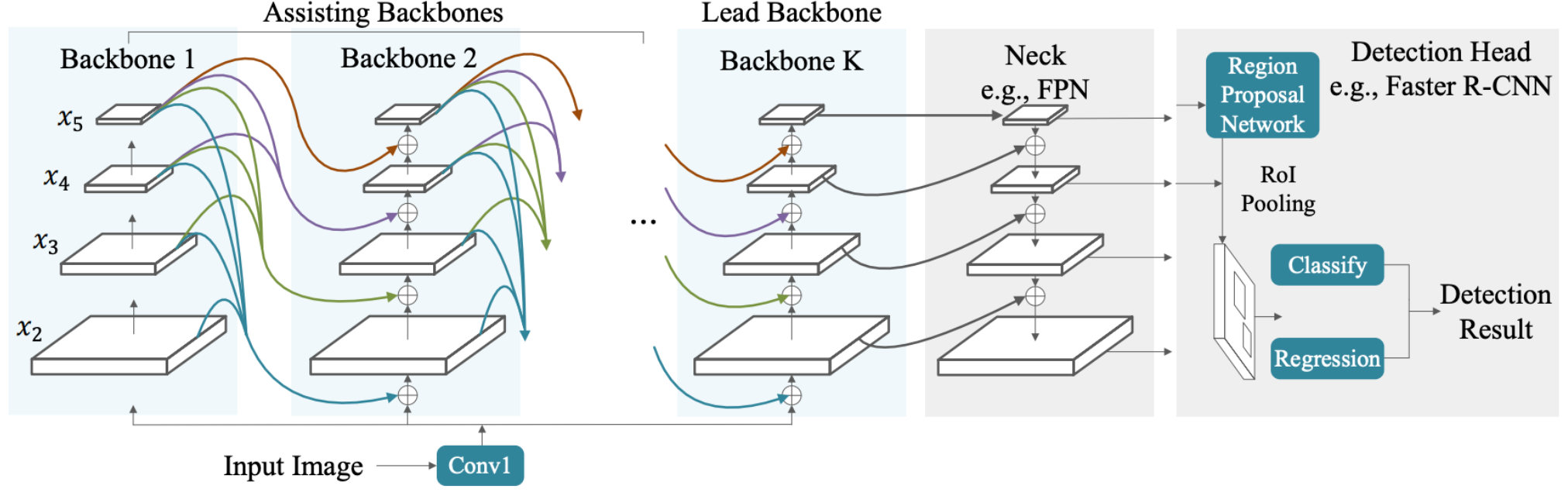}
    \caption{Instance segmentation model architecture and training pipeline of the method of Leng et al.}
    \label{fig:3_2_seg_a}
\end{figure}

Leng et al. demonstrate that a straightforward combination of well-proven methods can yield near-SoTA performance. The approach uses a Swin Transformer-Large~\cite{liu2021Swin} as the backbone, and the pipeline is based on CBNetV2~\cite{liang2021cbnetv2}, as shown in Figure~\ref{fig:3_2_seg_a}. In terms of data augmentations the method relies on a combination of AutoAugment~\cite{cubuk2018autoaugment}, ImgAug~\cite{imgaug} and Copy-Paste~\cite{ghiasi2021copypaste}.

\subsubsection{Shared second place - B}
Lu et al. make use of the popular HTC detector~\cite{Chen_2019_CVPR} with CBSwin-T~\cite{liang2021cbnetv2} backbone with CBFPN~\cite{liang2021cbnetv2} using group normalization, Mosaic, test-time augmentations and the Task-Specific Copy-Paste Data Augmentation Method~\cite{ghiasi2021copypaste} from a previous edition of the VIPriors Instance Segmentation challenge. Moreover, different backbones, namely ResNet~\cite{he2015deep}, ConvNeXt~\cite{liu2022convnet}, Swinv2~\cite{liu2021Swin} and CBNetv2~\cite{liang2021cbnetv2} are trained and combined using Model Soups~\cite{pmlr-v162-wortsman22a}. Multiple predictions are combined together using mask voting.

\subsubsection{Third place}

\begin{figure}
    \centering
    \includegraphics[width=\linewidth]{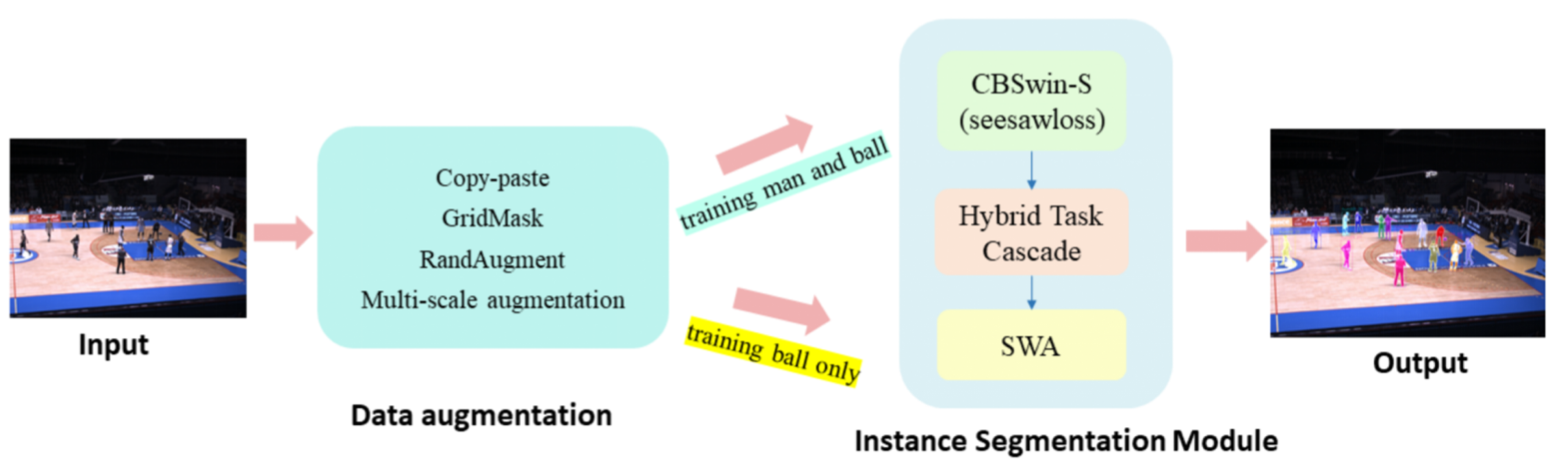}
    \caption{Overview of instance segmentation method by the third place competitor, Zhang et al.}
    \label{fig:3_3_seg}
\end{figure}

The method of Zhang et al. employs Location-aware MixUp, RandAugment, GridMask, Random scaling, Copy-paste, Multi-scale augmentation, and test-time augmentation in terms of data augmentation techniques. The model used is the popular HTC detector~\cite{Chen_2019_CVPR} and soft non-maxima suppression is applied on the predicted target boxes. The overall framework is depicted in figure~\ref{fig:3_3_seg}.

\subsubsection{Jury prize}

This year's jury prize is awarded to Sparse Instance Activation for Real-Time Instance Segmentation~\cite{Cheng2022SparseInst} by Cheng et al. The paper presents a method for instance segmentation using a novel representation of instance activation maps. These maps highlight informative regions for each object, which are then used to obtain instance-level features for recognition and segmentation. The method avoids the need for non-maximum suppression in post-processing by predicting objects in a one-to-one style using bipartite matching.
\subsection{Action Recognition}
Many of the popular Action Recognition models are deep networks that require a large amount of data for training, which can be challenging when there is limited data availability or insufficient compute resources. In line with the workshop's goals, we present the Kinetics400ViPriors dataset, which is a modified version of the well-known Kinetics400 dataset. We have created a smaller variant with 40k, 10k, and 20k clips for the train, validation, and test sets, respectively, while preserving the original number of action classes. Our aim is to motivate researchers in Action Recognition to develop efficient models that can leverage visual prior knowledge from the data.

For evaluation, we use the average classification accuracy across all classes on the test set. The accuracy for a single class is calculated as $ \mathrm{Acc} = \frac{P}{N} $, where $P$ represents the number of correct predictions for the evaluated class and $N$ is the total number of samples in that class. The average accuracy is determined by taking the mean of accuracies for all classes.

9 teams submitted solutions to the evaluation server, of which 3 teams submitted a report to qualify their submission to the challenge. The final rankings are shown in Table \ref{tab:actionrecognition}.

\begin{table*}[t]
\centering
\caption{Final rankings of the Action Recognition challenge. J indicates Jury prize.}
\renewcommand{\arraystretch}{2.0}
\begin{tabular}{@{}llc@{}}
\toprule
Ranking & Teams & Acc \\ \midrule
1 & \makecell[l]{
    \textbf{Xinran Song, Chengyuan Yang, Chang Liu, Yang Liu, Fang Liu, Licheng Jiao}\\
    \textbf{\textit{School of Artificial Intelligence, Xidian University, Xi’an, China.}}
} & \textbf{0.71} \\
2 & \makecell[l]{
    Wenxin He, Zihan Gao, Tianzhi Ma , Licheng Jiao\\
    \textit{School of Artificial Intelligence, Xidian University, Xi’an, China.}
} & 0.69 \\
3 \& J & \makecell[l]{
    Bo Tan, Yang Xiao, Wenzheng Zeng, Xingyu Tong, Zhiguo Cao, Joey Tianyi Zhou\\
    \textit{Huazhong University of Science and Technology (China) and CFAR (Singapore)}
} & 0.59 \\
\bottomrule
\end{tabular}
\label{tab:actionrecognition}
\end{table*}

\subsubsection{First place}
\begin{figure}
    \centering
    \includegraphics[width=\linewidth]{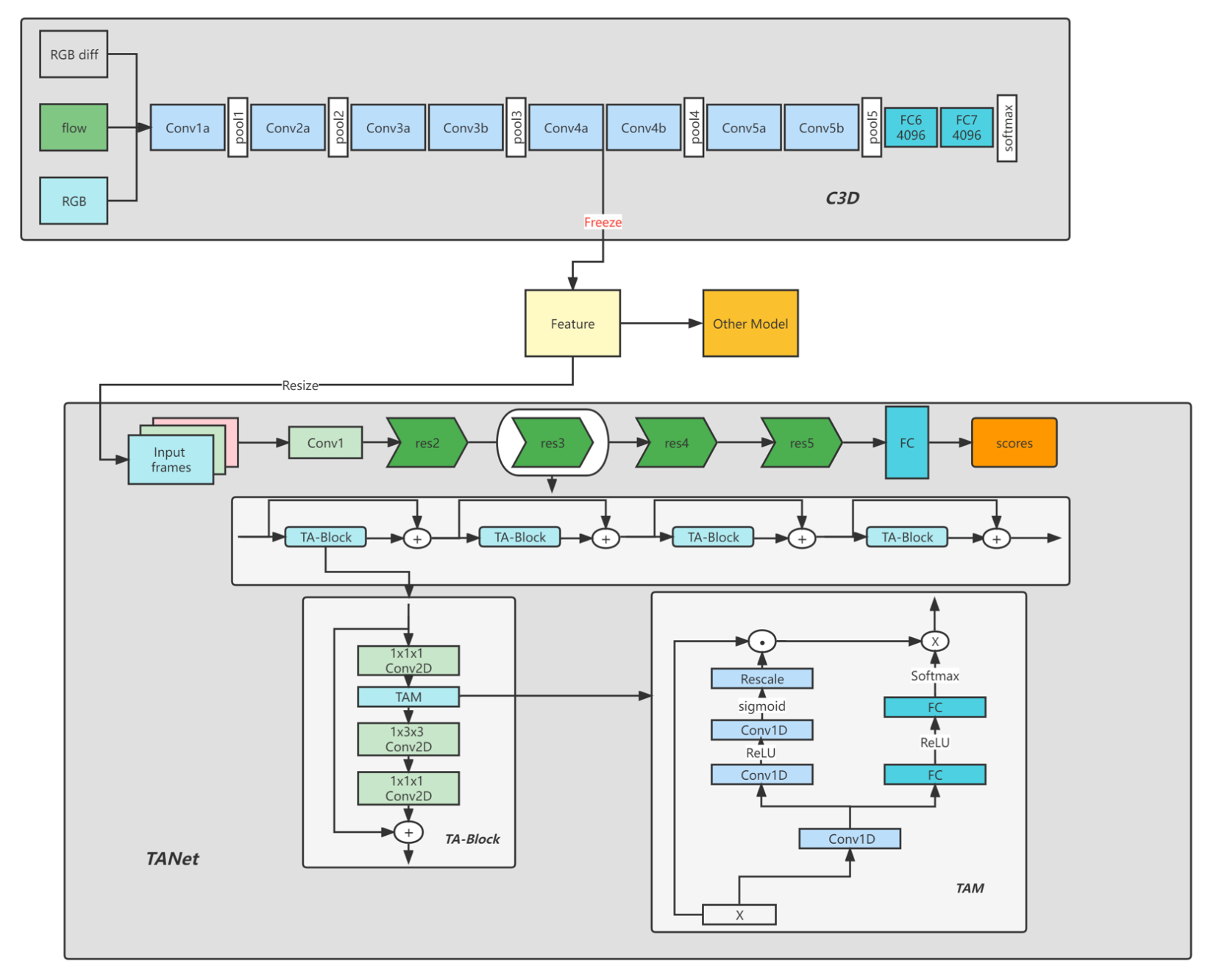}
    \caption{Method proposed by first place Song et al. from Xidian University.}
    \label{fig:arfirst}
\end{figure}

The authors train a selection of models, including R(2+1)D~\cite{Tran2018r21d}, SlowFast~\cite{Feichtenhofer2019sf}, CSN~\cite{csn}, X3D~\cite{x3d}, TANet~\cite{liu2019tanet} and Timesformer~\cite{timesformer}, and apply a model fusion approach by assigning different weights to the models and using the soft voting method to combine their results. In terms of data augmentation, frames were extracted from videos and a subset was selected by choosing every second frame. The videos were resized and noise was added through random flipping. During testing, TenCrop was used as a test-time enhancement. The evaluation involved ten-fold cross-validation, where the training dataset was combined with the validation dataset. An overview of the method is provided in Fig.~\ref{fig:arfirst}.

\subsubsection{Second place}
\begin{figure}
    \centering
    \includegraphics[width=\linewidth]{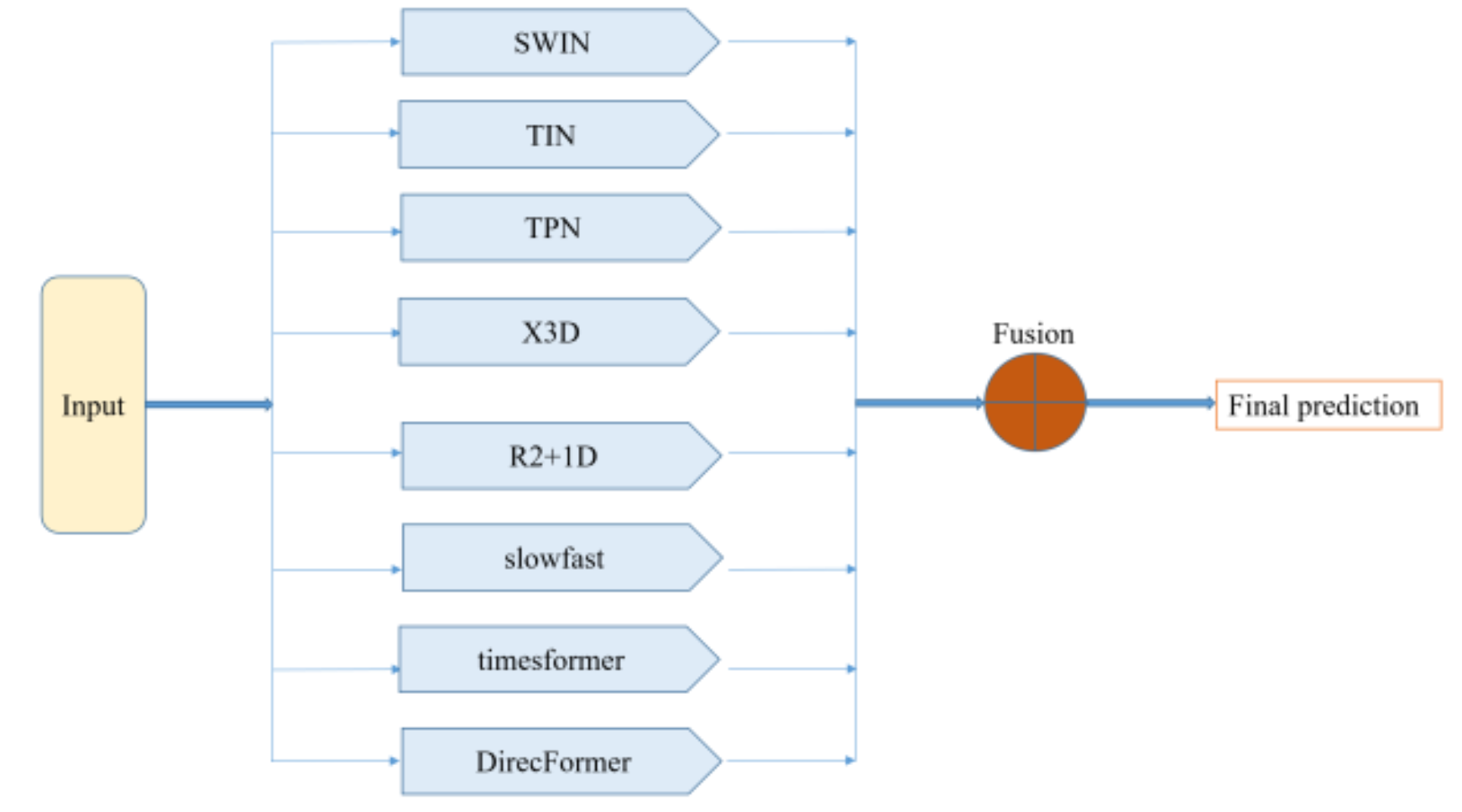}
    \caption{Schematic diagram of the method (inference mode) proposed by second place He et al. from Xidian University.}
    \label{fig:arsecond}
\end{figure}

The method proposes a multi-network dynamic fusion model combining a variety of backbones, including SlowFast~\cite{Feichtenhofer2019sf}, Timesformer~\cite{timesformer}, TIN~\cite{shao2020temporal}, TPN~\cite{tpn}, Video Swin Transformers~\cite{liu2021video}, R(2+1)D~\cite{Tran2018r21d}, X3D~\cite{x3d}, DirecFormer~\cite{truong2021direcformer}. Model predictions are combined as a weighted average by the prediction score of each model. Test-time augmentation with majority voting is used, as well as AutoAugment~\cite{cubuk2018autoaugment}, CutMix~\cite{yun2019cutmix}, and a variety of other spatial, photometric and temporal augmentations during training. An overview of the method is provided in Fig.~\ref{fig:arsecond}.

\subsubsection{Third place \& Jury prize}
The method combines self-supervised pre-training of various backbone models, optical flow estimation and model ensembling to train a data efficient video classification model. First, the 2D model encoders are pre-trained using the MoCo~\cite{he2020momentum} self-supervised representation learning framework on image data using the individual frames of the provided dataset. Next, optical flow features are extracted using the TVL-1~\cite{zach2017aduality} method. To correct for camera movement, consecutive image frames are aligned by calculcating the transformation matrix based on extracted SIFT features. Finally, a range of models including TSN~\cite{wang2016temporal}, TANet~\cite{liu2019tanet}, TPN~\cite{tpn}, SlowFast~\cite{Feichtenhofer2019sf}, CSN~\cite{csn} and Video MAE~\cite{tong2022videomae} are trained on the training data and pre-extracted optical flow features. MixUp~\cite{zhang2018mixup} and CutMix~\cite{yun2019cutmix} data augmentation is employed. Model ensembling is performed by concatenating the features of all models and training a single linear classifier layer after normalization. An ablation study is performed to show that self-supervised pre-training improves model performance.

\section{Conclusion}
We have summarized all solutions in Table~\ref{tab:conclusion} in terms of the encoder architecture, data augmentation techniques and main methods used.

Organizing the same challenges for the third year in a row gives a unique perspective on trends: which methods and/or architectures prevail over time, and which are replaced? The use of combining large numbers of models in ensembles and heavy data augmentation have been unchanging throughout the VIPiors challenge series. The models used in the ensembles are a mix of CNNs and Vision Transformers for the tasks of object detection and instance segmentation, whereas for image classification and action recognition Vision Transformers are not seeing use in our challenges. As for data augmentation, AutoAugment, MixUp and CutMix are unchanging constants in the training regimes of our competitors, regardless of the task.

Though we did not explicitly perform the required analysis, we cannot escape the impression that the simplicity of model ensembling is hard to beat with task-specific or domain-specific knowledge, especially when considering the effort required in design and implementation. Winning methods tend to use ensembling, and the bigger the ensemble, the better, as is shown in the ablation studies of some of the competitors' reports. If one is to follow this approach, we speculate that choosing models with a variety of inductive biases (e.g. CNNs and Vision Transformers) could make the ensemble more effective. However, such heavy use of ensembles may just be possible in our challenges because of the limited size of the datasets, which makes training many models feasible.

\Urlmuskip=0mu plus 1mu\relax
\bibliographystyle{splncs04}
\bibliography{main}

%




\end{document}